\documentclass[letterpaper]{article} 
\usepackage{aaai2027}  
\usepackage[hyphens]{url}  
\usepackage{graphicx} 
\usepackage{natbib}  
\usepackage{caption} 
\usepackage{amsmath}
\usepackage{booktabs}
\usepackage{array}
\newcolumntype{L}[1]{>{\raggedright\arraybackslash}p{#1}}
\newcolumntype{C}[1]{>{\centering\arraybackslash}p{#1}}
\newcommand{\actionname}[1]{\texttt{\detokenize{#1}}}
\nocopyright

\title{ProtoAct: Turning Wet-Lab Protocols into Embodied Robotic Actions}
\author{
Zhe~Liu\textsuperscript{\rm 1,\rm 2}\equalcontrib,
Jiaming~Gu\textsuperscript{\rm 1,\rm 2}\equalcontrib,
Zhaohui~Du\textsuperscript{\rm 1,\rm 2}\equalcontrib,
Zhe~Wang\textsuperscript{\rm 1,\rm 2}\corresponding,\\
Huanbo~Jin\textsuperscript{\rm 1,\rm 2},
Quan~Lu\textsuperscript{\rm 1,\rm 2},
Qi~Wang\textsuperscript{\rm 3},
Ting~Xiao\textsuperscript{\rm 1,\rm 2},\\
Minting~Pan\textsuperscript{\rm 4},
Dongzhan~Zhou\textsuperscript{\rm 4}
}
\affiliations{
\textsuperscript{\rm 1}Key Laboratory of Smart Manufacturing in Energy Chemical Process, Ministry of Education,\\
East China University of Science and Technology, Shanghai, China\\
\textsuperscript{\rm 2}Department of Computer Science and Engineering,\\
East China University of Science and Technology, Shanghai, China\\
\textsuperscript{\rm 3}Department of Laboratory Medicine, Ruijin Hospital,\\
Shanghai Jiao Tong University School of Medicine, Shanghai, China\\
\textsuperscript{\rm 4}AI for Science Center, Shanghai AI Laboratory, Shanghai, CN\\
wangzhe@ecust.edu.cn
}

\begin{document}

\maketitle

\begin{abstract}
Biological wet-lab protocols are written for trained researchers and often leave routine operations, state-dependent conditions, and contextual parameters implicit, making them difficult to translate into robot-executable actions. We present ProtoAct, a structured protocol-grounding framework that converts free-form biological procedures into state-aware, embodiment-ready action sequences. ProtoAct uses ProtoRAG to retrieve manually annotated examples for context-sensitive parsing, employs RefineChecker to detect and revise missing or inconsistent steps, and applies ActSchema to map the refined procedure into constrained JSON function sequences. We further introduce BioP2E, for which we manually annotate 22 cell-culture protocols into 258 monitoring conditions, 910 executable subtasks, and 962 grounded action calls. Evaluation across seven large language models demonstrates that ProtoAct can be effectively instantiated with different backbones. Ablations confirm that retrieval, posterior checking, and schema constraints make complementary contributions. The parsed subtasks further support demonstration collection and VLA model training, enabling successful execution in both simulation and real-robot settings. ProtoAct thus provides a practical interface between biological protocol understanding and embodied robotic execution.
\end{abstract}

\section{Introduction}

Biological wet-lab experiments are commonly guided by protocols, namely written instructions describing how samples are prepared, processed, monitored, and measured. As automated laboratories increasingly use robotic systems to improve experimental efficiency and reproducibility, these protocols provide a natural source of task specifications. However, they are written for trained researchers and are not directly machine-executable: routine operations may be omitted, equivalent actions may be described differently, and parameters often depend on context. Converting protocols into structured, robot-readable action representations is therefore essential for connecting experimental knowledge with automated execution.

Recent studies have applied large language models to protocol understanding, generation, and planning. BioPlanner converts biological protocols into pseudocode and evaluates the planning ability of large language models using optional pseudo-functions \cite{odonoghue2023bioplanner}. ProtoMed-LLM adopts a predefined set of experimental actions for pseudocode extraction and model evaluation \cite{yi2025protomed}. BioMARS integrates protocol generation, pseudocode conversion, and multimodal anomaly detection within a multiagent robotic system \cite{qiu2025biomars}. These studies show that language models can organize experimental instructions into structured forms. However, pseudocode and free-form plans do not fully specify what embodied execution requires: which physical action to perform, when to trigger it, which parameters to use, and how each step depends on the experimental state.

This gap is especially important for vision-language-action models, which learn robot behavior from visual observations, language instructions, and demonstrations. Equivalent operations may be described differently across protocols, while one sentence may combine several physical actions. Reliable execution thus requires an intermediate representation that preserves protocol meaning while making monitoring conditions, action primitives, parameters, and procedural sequences explicit. It should remain readable for human inspection and also support automatic evaluation, demonstration collection, and policy learning.

To address this need, we introduce ProtoAct, a structured protocol-grounding framework that translates biological procedures into action programs designed for embodied execution. ProtoAct generates monitoring conditions and subtask sequences, checks them against experiment-specific requirements, and grounds the refined procedure in a parameterized action schema. We further construct BioP2E, a manually annotated cell-culture protocol dataset aligned with these embodied task representations. Finally, we validate ProtoAct outputs through robotic task construction, demonstration collection, and VLA training. Experiments in simulation and on a physical robot show their utility for embodied data generation and policy execution. The code is available at \url{https://github.com/gjm112233/ProtoAct}. Our contributions are summarized as follows.

\begin{itemize}
    \item We introduce, to our knowledge, the first protocol-grounding formulation for embodied wet-lab execution, connecting free-form experimental instructions with state-aware task representations and parameterized robotic actions.
    \item We construct BioP2E by manually annotating 22 cell-culture protocols into 258 monitoring conditions, 910 executable subtasks, and 962 grounded action calls, providing a resource for evaluation and embodied task construction.
    \item We establish a multi-level benchmark for biological protocol grounding and further validate the grounded representations through VLA training and execution in both simulation and real-robot settings.
\end{itemize}

\begin{figure*}[t]
    \centering
    \includegraphics[width=0.98\textwidth]{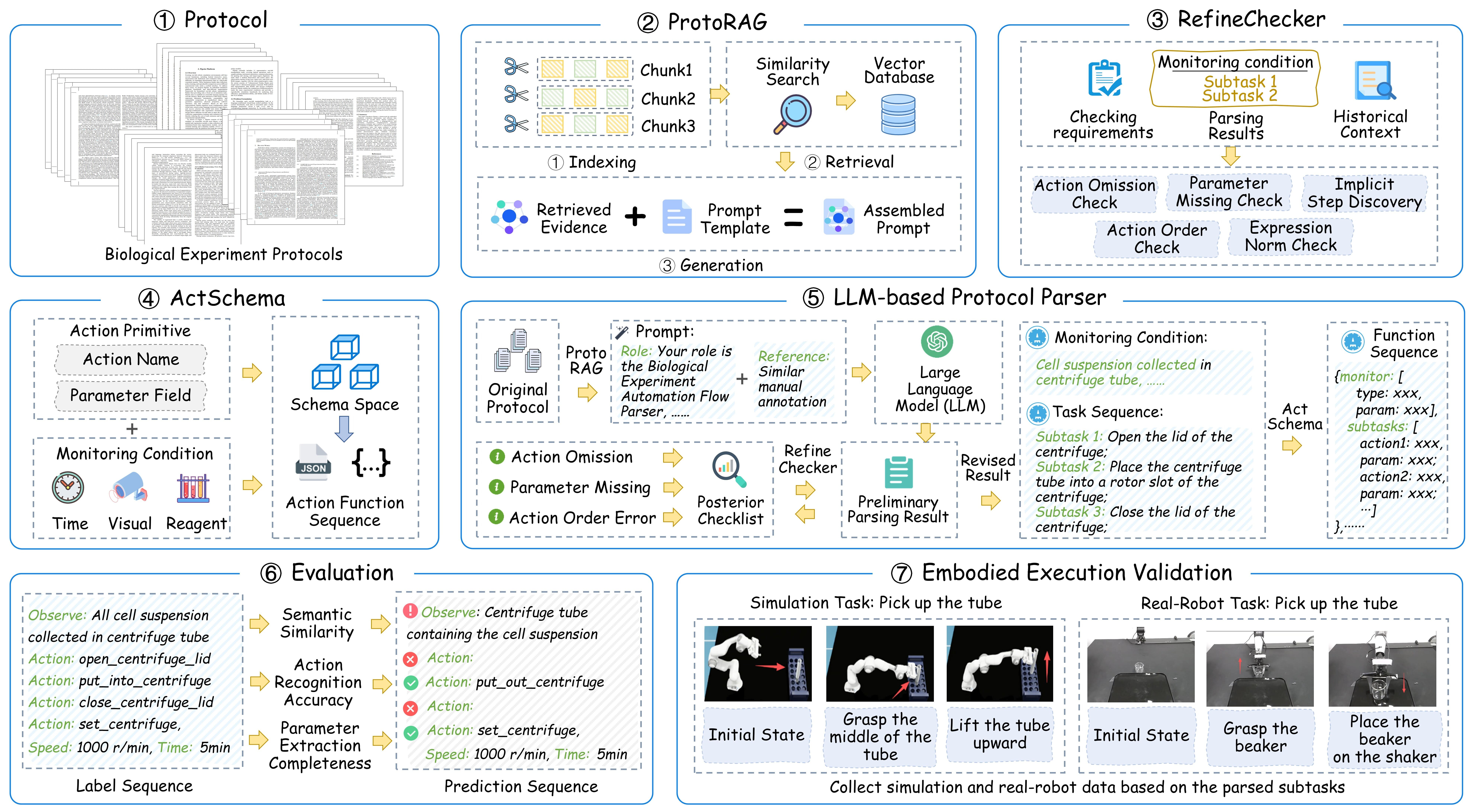}
    \caption{Architecture of the ProtoAct system.}
    \label{fig:protoact}
\end{figure*}

\section{Related Work}

\subsection{Biological Protocol Parsing and Laboratory Automation}

Biological experimental protocols typically describe experimental procedures in natural language. Recent studies have explored the use of large language models and agent-based systems for protocol understanding, generation, and structured transformation. BioPlanner formulates protocol generation as pseudo-function selection and stepwise pseudocode reconstruction \cite{odonoghue2023bioplanner}. ProtoMed-LLM evaluates the protocol generation capabilities of large language models using a predefined set of experimental actions \cite{yi2025protomed}. ProtoCode converts PCR procedures into machine-readable representations \cite{jiang2024protocode}. 
For experimental execution, RoboCulture combines robotic manipulation, visual feedback, and behavior trees to automate long-duration biological culture workflows \cite{angers2025roboculture}. BioProVLA-Agent integrates protocol parsing, visual state verification, and VLA models to support closed-loop execution in wet-lab environments \cite{du2026bioprovla}. However, existing studies still lack an intermediate representation that connects protocol parsing with robotic execution. To address this gap, ProtoAct employs a two-stage parsing pipeline that first produces task decompositions for human interpretation and then converts them into structured action representations for robotic execution.

\subsection{Action Space Modeling for Robot Execution}

Mapping natural language instructions to executable action sequences is a key step in connecting large language models with robotic systems. SayCan combines language-model-based semantic scores with robotic skill affordances to select skills that are both semantically appropriate and executable \cite{ahn2022saycan}. Code as Policies generates programs that invoke control APIs, whereas ProgPrompt explicitly provides the available action set to guide task planning under environmental constraints \cite{liang2023codeaspolicies,singh2023progprompt}. ProtoMed-LLM uses a predefined set of experimental actions to standardize the generation and evaluation of protocol pseudo-functions \cite{yi2025protomed}. However, existing methods provide limited modeling of state dependencies and procedural constraints in biological wet-lab experiments. To address this limitation, ProtoAct uses monitoring conditions and subtask sequences as intermediate representations. ActSchema further defines monitoring primitives, action primitives, and their parameter schemas, enabling these representations to be mapped into \texttt{JSON} action function sequences for automated evaluation and robotic task construction.

\subsection{Robotic Manipulation Learning and Simulation Verification Based on VLA Models}
Vision-language-action models jointly process visual observations, language instructions, and robot states to generate language-conditioned robot actions. RT-2 improves generalization by jointly fine-tuning a pretrained vision-language model on robotic trajectory data \cite{brohan2023rt2}. OpenVLA provides a general-purpose VLA model and fine-tuning framework trained on large-scale robotic demonstrations \cite{kim2024openvla}. \(\pi_0\) adopts a flow-matching architecture to learn general manipulation policies \cite{black2024pi0}, while SmolVLA focuses on lightweight and cost-effective deployment \cite{shukor2025smolvla}. 
In addition, LIBERO provides a simulation benchmark for lifelong robot learning and knowledge transfer \cite{liu2023libero}, and AutoBio develops a digital simulation platform for biological laboratory automation \cite{lan2025autobio}. Existing studies have paid limited attention to connecting protocol parsing results with robot learning tasks. This work uses VLA models to validate that the subtasks generated by ProtoAct can support data collection, model training, and execution evaluation in both simulated and real-world robotic tasks, thereby establishing a complete workflow from protocol parsing to embodied execution validation.

\section{ProtoAct Pipeline}

\subsection{Overview}

ProtoAct is designed to bridge biological protocol understanding and embodied robotic execution by converting natural-language procedures into structured and verifiable workflow representations at both the subtask and action-function levels, as shown in Figure~\ref{fig:protoact}. Given a raw biological experimental protocol as input, the system first combines a large language model with ProtoRAG to parse the protocol into natural-language monitoring conditions and subtask sequences. The monitoring conditions specify the requirements that must be satisfied before subsequent subtasks can be executed, while each subtask represents the smallest physical action that the robot needs to perform under the corresponding condition. This representation preserves the temporal logic among experimental steps and provides clear boundaries for subsequent action mapping.

After obtaining the initial parsing result, ProtoAct applies RefineChecker as a posterior checking mechanism to verify and revise the preliminary result according to experiment-specific notes provided by the user. This process preserves the correctly parsed content while incorporating domain knowledge to address the large language model's incomplete understanding of experimental logic. Finally, under the constraints of ActSchema, the monitoring conditions and subtask sequences are mapped into structured JSON action function sequences. Compared with directly generating low-level control commands from raw experimental protocols, ProtoAct introduces an intermediate representation between semantic parsing and action function mapping. This design makes the parsing results easier for users to inspect and revise, while also supporting the subsequent conversion into action function sequences with a unified format.

To support knowledge base uploading, protocol input, result display, and multi turn interaction, the system uses Streamlit to build the interactive interface. LangChain is used to organize retrieval, prompt construction, model invocation, and dialogue history management. 

\subsection{ProtoRAG: Retrieval-Augmented Protocol Parsing}

\begin{figure*}[t]
    \centering
    \includegraphics[width=0.98\textwidth]{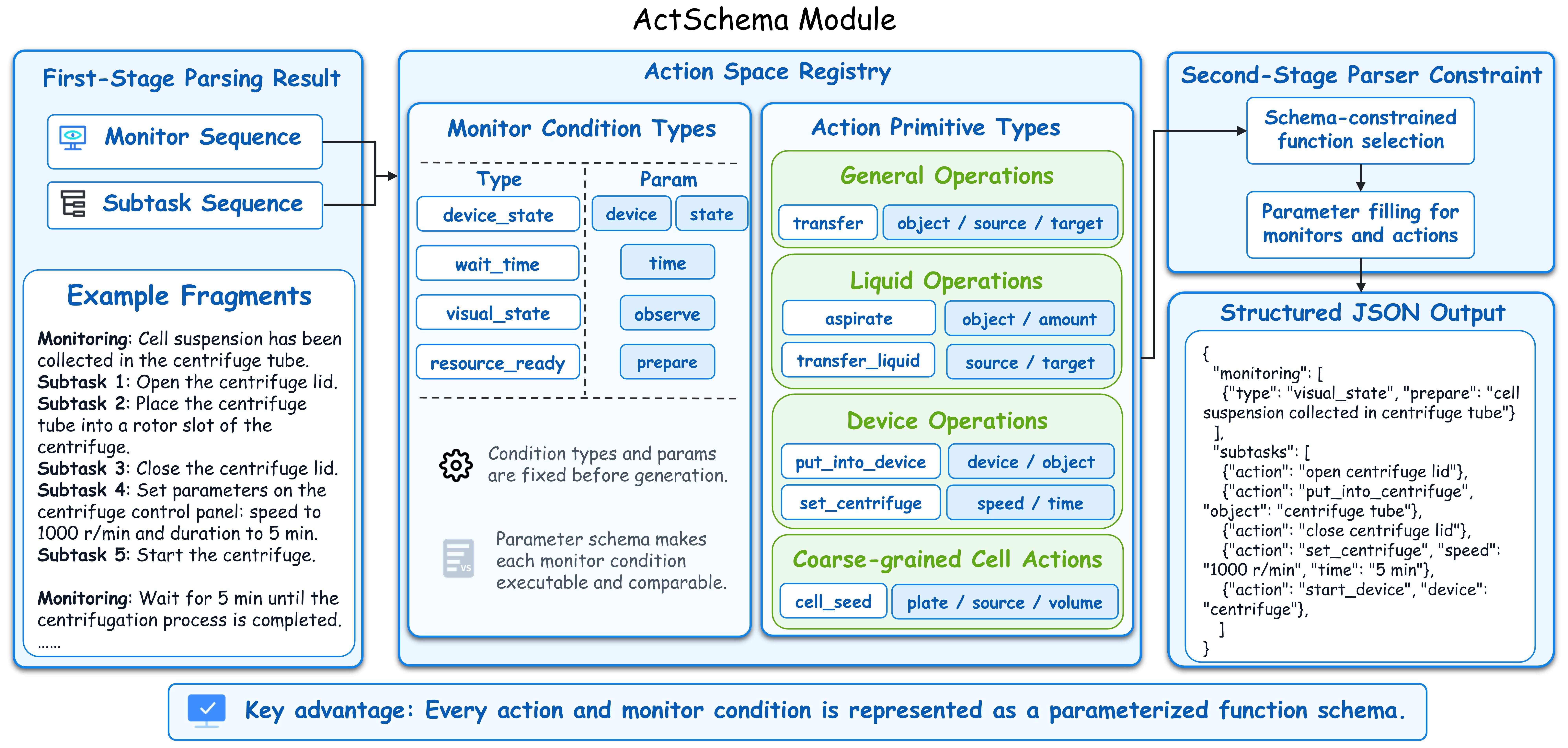}
    \caption{Architecture of the ActSchema module.}
    \label{fig:actschema}
\end{figure*}

ProtoRAG is used to improve the consistency and completeness of the parsing results. A workflow illustration of ProtoRAG is provided in the supplementary material. 
ProtoRAG consists of offline knowledge construction and online retrieval-augmented generation. In the offline stage, manually annotated protocol examples are used as the content of the knowledge base. Each example consists of a raw protocol fragment and its corresponding standardized parsing result. The system first splits the knowledge base text into chunks, then uses an embedding model to encode the text chunks into vectors, which are stored in a local Chroma vector database. 
In the online parsing stage, the new experimental protocol is used as the query text for the vector retriever. The system computes the semantic similarity between the query text and all text chunks in the knowledge base, and returns the top \(k\) document fragments with the highest similarity:

\begin{equation}
R_k(P) = \operatorname{TopK}\left(\operatorname{sim}(E(P), v_j), k\right).
\label{eq:retrieval}
\end{equation}

Here, \(R_k(P)\) denotes the top \(k\) reference fragments retrieved for the input protocol \(P\), \(\operatorname{sim}\) denotes the vector similarity function, \(E\) denotes the embedding model, \(v_j\) denotes the vector representation of the \(j\)-th text chunk, and \(\operatorname{TopK}\) denotes the operation of selecting the \(k\) results with the highest similarity.

The retrieved results are then formatted as reference materials and combined with the role definition, action granularity rules, and other instructions to form the final prompt. Let \(G\) denote the set of parsing rules. The parsing process of ProtoRAG can be represented as:

\begin{equation}
Y^{(0)} = f_{\mathrm{ProtoRAG}}\left(P, R_k(P), G\right).
\label{eq:protorag}
\end{equation}

Here, \(Y^{(0)}\) denotes the initial monitoring conditions and subtask sequences. \(P\) denotes the input raw protocol.

The main role of ProtoRAG is to provide retrievable external priors for protocol parsing. By dynamically retrieving similar cases according to the input, ProtoRAG offers task-relevant references for different experimental procedures and helps improve the quality of the initial parsing result.

\subsection{RefineChecker: Post-parsing Verification and Targeted Revision}

RefineChecker is used to verify and revise the initial natural language parsing result. An architecture diagram for RefineChecker is included in the supplementary material.
The input of RefineChecker consists of two parts: the initial parsing result \(Y^{(0)}\), and the checking information \(U\) provided by the user for the current experiment. RefineChecker compares \(Y^{(0)}\) with \(U\). If the relevant notes have already been covered, the result is kept unchanged. If omissions or errors are identified, the corresponding fragments are reanalyzed, and revised monitoring conditions and subtask sequences are generated.

\begin{figure}[t]
    \centering
    \includegraphics[width=\columnwidth]{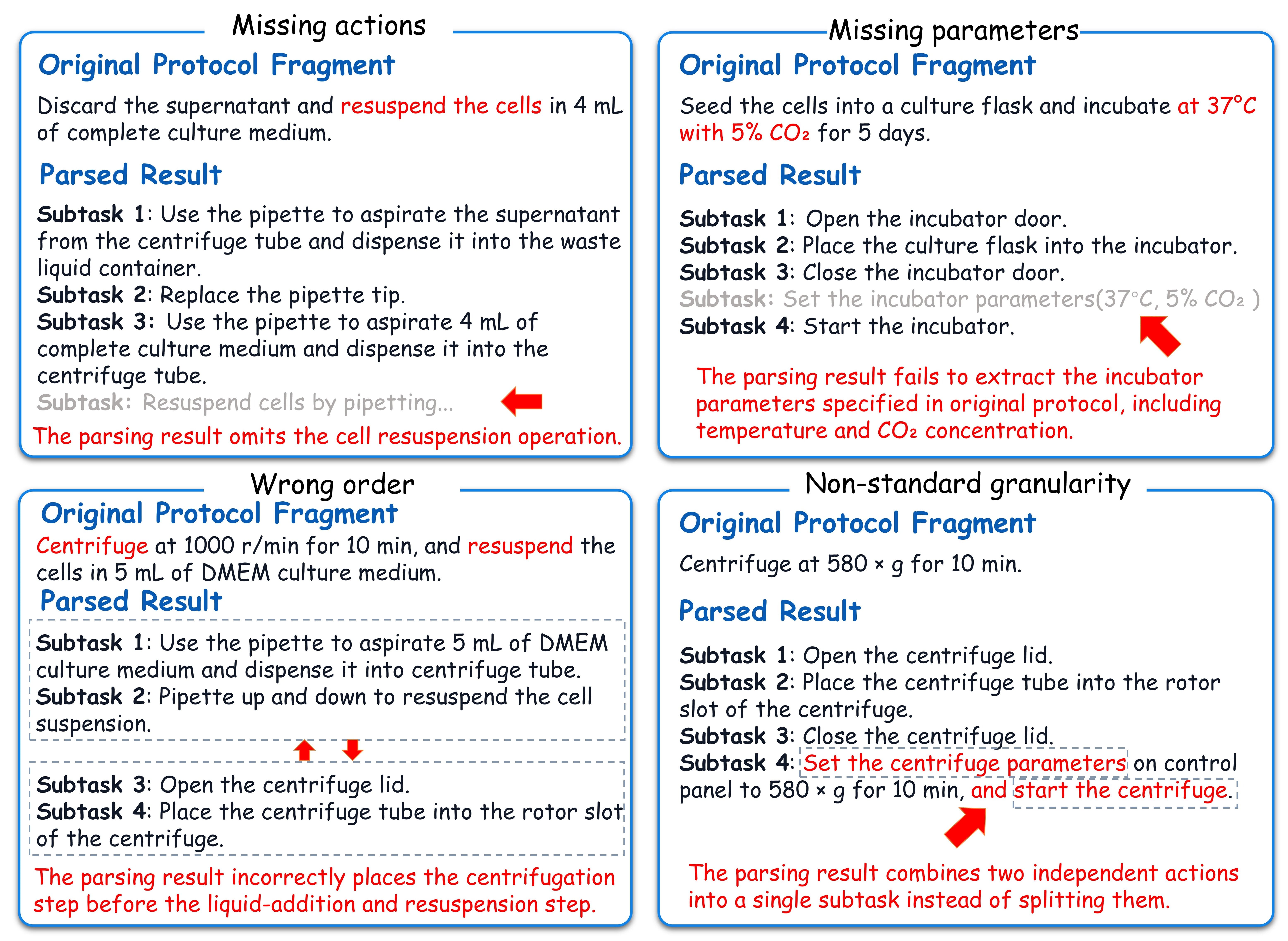}
    \caption{Cases of four types of parsing defects.}
    \label{fig:failure_cases}
\end{figure}

RefineChecker focuses on four types of parsing defects, as shown in Figure~\ref{fig:failure_cases}. The first type is missing actions, where implicit operations in the raw protocol are absent from the subtask sequences. The second type is missing parameters, where parameter information in the raw protocol is not extracted. The third type is incorrect order, where the parsed result violates the dependency between experimental operations. The fourth type is improper parsing granularity, such as merging multiple consecutive physical actions into a single subtask. The verification process of RefineChecker can be formalized as:

\begin{equation}
Y^{(1)} = f_{\mathrm{RefineChecker}}\left(Y^{(0)}, U\right).
\label{eq:refinechecker}
\end{equation}

Here, \(Y^{(1)}\) denotes the intermediate representation after post-parsing verification and revision.

To support continuous comparison among the initial parsing result, user provided information, and revised output, ProtoAct maintains dialogue history in the LangChain workflow, allowing the model to access the initial parsing result. This mechanism is lightweight and interactive, making it suitable for forming a feedback loop between human review and model based parsing.

\subsection{ActSchema: Constrained Function Selection and Parameter Filling}

ActSchema is used to convert the revised intermediate representation \(Y^{(1)}\) into a structured action function sequence. Its structure is shown in Figure~\ref{fig:actschema}. Although \(Y^{(1)}\) is readable and suitable for human inspection, it is still expressed in natural language, making it difficult to use directly for automatic evaluation. To convert these natural language fragments into function calls recognizable by the system, we constructs ActSchema to constrain the process of function selection and parameter filling.

ActSchema defines monitoring condition types, action primitive names, and the required parameter fields. Based on these schemas, ActSchema maps the natural language monitoring conditions and subtask sequences into structured function calls. Let \(z_t\) denote the \(t\)-th text unit extracted from \(Y^{(1)}\), which contains the monitoring condition \(m_t\) and several subtasks \(s_{t,j}\). The mapping process of ActSchema can be represented as:

\begin{equation}
a_t = f_{\mathrm{ActSchema}}(z_t, \Omega).
\label{eq:actschema_mapping}
\end{equation}

Here, \(a_t\) denotes the structured function call mapped from the text unit \(z_t\), and \(\Omega\) denotes the complete function schema space defined by ActSchema.

All function calls are then organized according to the execution order of monitoring conditions and subtasks to form the action function sequence:

\begin{equation}
A = \{a_t\}_{t=1}^{N}.
\label{eq:action_sequence}
\end{equation}

With ActSchema constraints, ProtoAct formulates open-ended language generation as constrained function selection and parameter filling, facilitating output alignment and automatic evaluation.

To ensure that constraint information can be consistently used during parsing, the system organizes the complete monitoring condition and action primitive schemas into external configuration files, which are provided to the model in the parsing workflow. 
ActSchema serves as the bridge from natural language task descriptions to standardized function representations in the overall pipeline.

\section{Experiment}

\subsection{BioP2E Dataset}

To evaluate structured protocol parsing and its applicability to embodied robotic tasks, we manually annotate 22 public wet-lab cell culture protocols collected from Protocols.io~\cite{teytelman2016protocols} to construct BioP2E. Each protocol is annotated at two levels: natural-language monitoring conditions and subtask sequences at the semantic level, and schema-constrained JSON action function sequences at the execution level. The detailed annotation procedure is provided in the supplementary material.
All annotations are manually reviewed by experts with biological experiment backgrounds for experimental validity, semantic completeness, and executability. Among the 22 annotated protocols, two are reserved as reference protocols for constructing the ProtoRAG knowledge base. The remaining 20 protocols are used exclusively for model evaluation, ensuring that the retrieval knowledge base does not overlap with the evaluation set. Dataset statistics are summarized in Table~\ref{tab:biop2e_stats}.

\begin{table}[t]
\centering
\setlength{\tabcolsep}{4pt}
\begin{tabular}{lr}
\toprule
Statistic & Value \\
\midrule
Biological experiment protocols & 22 \\
Raw protocol steps & 103 \\
Monitoring primitive types & 4 \\
Action primitives & 30 \\
Monitoring conditions & 258 \\
Subtasks & 910 \\
Action functions & 962 \\
Average subtasks per protocol & 41.36 \\
Average action calls per protocol & 43.73 \\
\bottomrule
\end{tabular}
\caption{Overall statistics of the BioP2E dataset.}
\label{tab:biop2e_stats}
\end{table}

\subsection{Evaluation Metrics}

We evaluate biological protocol parsing by comparing model outputs with manually annotated ground truth at three levels: action primitive and monitoring condition recognition, parameter extraction and value similarity, and action sequence order consistency. A GPT-based evaluator is further used to assess the overall semantic validity and executability of the parsing results.

\subsubsection{Action Primitives and Monitoring Condition Recognition Evaluation}

For each protocol, the predicted action function sequence is matched against the manually annotated ground-truth sequence. We count true positives (TP), false positives (FP), and false negatives (FN), and report the F1 score for both action primitives and monitoring condition types.

\subsubsection{Parameter Field Extraction and Parameter Value Similarity Evaluation}

Parameter fields are evaluated using F1 under the same matching procedure. Parameter values are evaluated using BLEU~\cite{papineni2002bleu} and SciBERTScore~\cite{beltagy2019scibert}, which measure lexical overlap and semantic similarity, respectively.

BLEU measures the n-gram overlap between predicted and ground-truth parameter values:

\begin{equation}
\mathrm{BLEU}=\exp\left(\sum_{n=1}^{N} w_n \log p_n\right).
\label{eq:bleu}
\end{equation}

Here, \(p_n\) denotes the matching proportion of continuous token fragments of length \(n\) between the two values, \(w_n\) denotes the weight assigned to each \(n\)-gram order, and \(N\) denotes the maximum \(n\)-gram order.

Semantically equivalent parameter values may have different surface forms, we further compute their embedding similarity using SciBERT:

\begin{equation}
\mathrm{SciBERTScore}=\frac{1}{N}\sum_{i=1}^{N}
\frac{\left\langle E(a_i^{\mathrm{pred}}), E(a_i^{\mathrm{GT}})\right\rangle}
{\lVert E(a_i^{\mathrm{pred}})\rVert
\lVert E(a_i^{\mathrm{GT}})\rVert}.
\label{eq:scibertscore}
\end{equation}

Here, \(E\) denotes the SciBERT encoder, \(a_i^{\mathrm{pred}}\) and \(a_i^{\mathrm{GT}}\) denote the predicted and ground-truth parameter values, and \(N\) denotes the number of evaluated parameters.

\subsubsection{Action Sequence Order Consistency Evaluation}

We use the normalized Levenshtein distance \(L_{\mathrm{dn}}\), to evaluate the order consistency between predicted and ground-truth action sequences~\cite{levenshtein1966binary}. Given the predicted action sequence \(S_{\mathrm{pred}}\) and the ground-truth sequence \(S_{\mathrm{GT}}\), it is defined as:

\begin{equation}
L_{\mathrm{dn}}=
\frac{L_d(S_{\mathrm{pred}}, S_{\mathrm{GT}})}
{|S_{\mathrm{GT}}|}.
\label{eq:ldn}
\end{equation}

Here, \(L_d\) denotes the minimum number of insertions, deletions, and substitutions required to transform \(S_{\mathrm{pred}}\) into \(S_{\mathrm{GT}}\), and \(|S_{\mathrm{GT}}|\) denotes the length of the ground truth action sequence. A lower \(L_{\mathrm{dn}}\) indicates better sequence consistency.

\subsubsection{Parsing Quality Evaluation with a GPT Evaluator}

Following prior LLM-based evaluation studies~\cite{liu2023geval}, we use GPT-5.3 to complement the automatic metrics. The evaluator receives the original protocol, the generated monitoring conditions and subtask sequences, and the predefined parsing rules, to assess the overall semantic validity and executability of the parsing result.
An example GPT-evaluator output is included in the supplementary material. This evaluation helps identify experimental validity and potential execution risks that may not be captured by automatic metrics.

\subsection{Multi-Model Comparison on Protocol Parsing}

To compare the performance of different large language models on biological protocol parsing, we evaluate seven models on 20 cell culture protocols. The evaluation is conducted from two aspects: monitor condition and action primitive. 

\begin{table*}[t]
\centering
{\small
\setlength{\tabcolsep}{1mm}
\begin{tabular}{lccccccccc}
\toprule
Models & Monitor Type & \multicolumn{3}{c}{Monitor Param.} & Action Prim. & \multicolumn{3}{c}{Action Param.} & Action Order \\
\cmidrule(lr){2-2}\cmidrule(lr){3-5}\cmidrule(lr){6-6}\cmidrule(lr){7-9}\cmidrule(lr){10-10}
 & F1 & F1 & BLEU & SciBERT & F1 & F1 & BLEU & SciBERT & \(L_{\mathrm{dn}}\) \\
\midrule
qwen3-max & 0.88{\(\pm\)}0.06 & 0.89{\(\pm\)}0.05 & \underline{0.74{\(\pm\)}0.12} & \underline{0.95{\(\pm\)}0.03} & 0.93{\(\pm\)}0.06 & 0.94{\(\pm\)}0.05 & \textbf{0.87{\(\pm\)}0.07} & \textbf{0.98{\(\pm\)}0.01} & 0.13{\(\pm\)}0.10 \\
qwen-max & 0.88{\(\pm\)}0.11 & 0.89{\(\pm\)}0.10 & {0.73{\(\pm\)}0.11} & {0.93{\(\pm\)}0.03} & 0.92{\(\pm\)}0.05 & 0.93{\(\pm\)}0.04 & 0.81{\(\pm\)}0.11 & {0.94{\(\pm\)}0.02} & 0.17{\(\pm\)}0.10 \\
gpt-5.3 & \underline{0.89{\(\pm\)}0.06} & {0.90{\(\pm\)}0.06} & \textbf{0.77{\(\pm\)}0.09} & \textbf{0.96{\(\pm\)}0.02} & \textbf{0.95{\(\pm\)}0.04} & \textbf{0.96{\(\pm\)}0.03} & \underline{0.85{\(\pm\)}0.07} & \underline{0.97{\(\pm\)}0.01} & \textbf{0.09{\(\pm\)}0.10} \\
deepseek-v3.2 & \textbf{0.90{\(\pm\)}0.07} & \textbf{0.92{\(\pm\)}0.06} & 0.71{\(\pm\)}0.11 & 0.94{\(\pm\)}0.02 & {0.93{\(\pm\)}0.05} & \underline{0.95{\(\pm\)}0.04} & 0.84{\(\pm\)}0.09 & {0.95{\(\pm\)}0.02} & 0.14{\(\pm\)}0.11 \\
deepseek-r1 & \textbf{0.90{\(\pm\)}0.07} & \underline{0.91{\(\pm\)}0.07} & 0.71{\(\pm\)}0.10 & 0.94{\(\pm\)}0.02 & \underline{0.94{\(\pm\)}0.05} & 0.94{\(\pm\)}0.05 & 0.84{\(\pm\)}0.07 & {0.94{\(\pm\)}0.02} & 0.13{\(\pm\)}0.10 \\
kimi-k2 & 0.88{\(\pm\)}0.08 & 0.89{\(\pm\)}0.08 & 0.72{\(\pm\)}0.13 & {0.92{\(\pm\)}0.03} & {0.92{\(\pm\)}0.06} & {0.91{\(\pm\)}0.05} & 0.82{\(\pm\)}0.11 & {0.93{\(\pm\)}0.01} & \underline{0.11{\(\pm\)}0.10} \\
glm-4.5 & \textbf{0.90{\(\pm\)}0.08} & {0.90{\(\pm\)}0.07} & 0.70{\(\pm\)}0.11 & 0.94{\(\pm\)}0.02 & \underline{0.94{\(\pm\)}0.05} & {0.93{\(\pm\)}0.04} & 0.84{\(\pm\)}0.07 & {0.96{\(\pm\)}0.01} & \underline{0.11{\(\pm\)}0.09} \\
\bottomrule
\end{tabular}
}
\caption{Comparison of protocol parsing performance across monitor conditions and action primitives. Each cell reports the mean and standard deviation of the corresponding metric. The best and second-best results are marked in bold and underlined.}
\label{tab:monitor_comparison}
\label{tab:action_comparison}
\end{table*}

As shown in Table~\ref{tab:monitor_comparison}, the F1 scores of different models are generally close in monitor condition type recognition. Among them, deepseek-r1, glm-4.5, and deepseek-v3.2 perform relatively better, indicating that they can more effectively identify monitor conditions from the original protocols and assign them to the correct types. For parameter parsing, deepseek-v3.2 performs better in monitor condition parameter field extraction, while gpt-5.3 achieves better similarity for monitor parameter values. For action parsing, gpt-5.3 achieves the best results in action primitive F1, action parameter F1, and \(L_{\mathrm{dn}}\), suggesting that its outputs are closer to the manual annotations in terms of action coverage, parameter field extraction, and order consistency. Qwen3-max performs best in action parameter value similarity, indicating that its generated parameter values are closer to the manual annotations in both lexical form and semantic meaning.

\subsection{Ablation Experiments}

To analyze the contribution of each module in the ProtoAct pipeline, we conduct ablation experiments on ProtoRAG, RefineChecker, and ActSchema using Qwen3-Max as the base model. Qwen3-Max is selected for its strong overall performance in the multi-model comparison. The complete ablation results are reported in the supplementary material.

The full model achieves the best results in monitor condition parsing, action primitive recognition, parameter extraction, and sequence consistency, indicating that the three modules all contribute to the overall parsing performance. Specifically, removing ProtoRAG leads to the most obvious decrease in monitor condition-related metrics, suggesting that reference examples help the model identify implicit monitor conditions and their parameters that depend on experimental context. Removing ActSchema causes the largest drop in action primitive F1 and action parameter F1, showing that the predefined action schema space can constrain action function generation and improve the regularity of parameter structures. Removing RefineChecker increases omissions in monitor conditions and action primitives and raises \(L_{\mathrm{dn}}\), indicating that this module can supplement missing content through posterior checking and help maintain a reasonable experimental procedure order.

\subsection{Comprehensive Comparison of Parsing Quality Based on the GPT Evaluator}

We use GPT-5.3 to evaluate the monitoring conditions and subtask sequences generated by different models in the first parsing stage and to compare the effects of ProtoRAG and RefineChecker on parsing quality.

The GPT evaluation results reported in the supplementary material show that introducing RefineChecker improves the scores of most models, indicating that posterior checking helps correct action omissions and procedural ordering errors in the parsing results. In the vascular endothelial cell protocol, Qwen3-Max with RefineChecker recovers the implicit transfer operation before centrifugation but misses the removal of the washing solution, resulting in a score of 4. With ProtoRAG, the model uses annotated examples to complete the washing procedure and achieves a score of 5. Overall, ProtoRAG and RefineChecker provide complementary improvements in parsing completeness and consistency.

\subsection{Simulation and Real-Robot Validation for Embodied Execution}

\begin{table}[t]
\centering
\setlength{\tabcolsep}{4pt}
\begin{tabular}{p{0.42\columnwidth}ccc}
\toprule
Task & \begin{tabular}[c]{@{}c@{}}Episodes\end{tabular} & \begin{tabular}[c]{@{}c@{}}\(\pi_0\) \end{tabular} & \begin{tabular}[c]{@{}c@{}}SmolVLA \end{tabular} \\
\midrule
Tube pickup & 100 & 59 & 95 \\
Centrifuge lid opening & 100 & 90 & 96 \\
Centrifuge lid closing & 100 & 96 & 98 \\
Petri dish take-out & 100 & 34 & 66 \\
Petri dish placement & 100 & 52 & 96 \\
\bottomrule
\end{tabular}
\caption{Execution success rates of two VLA models on five simulation tasks.}
\label{tab:vla_success}
\end{table}

\subsubsection{Simulation Dataset}

\begin{figure*}[t]
    \centering
    \includegraphics[width=0.98\textwidth]{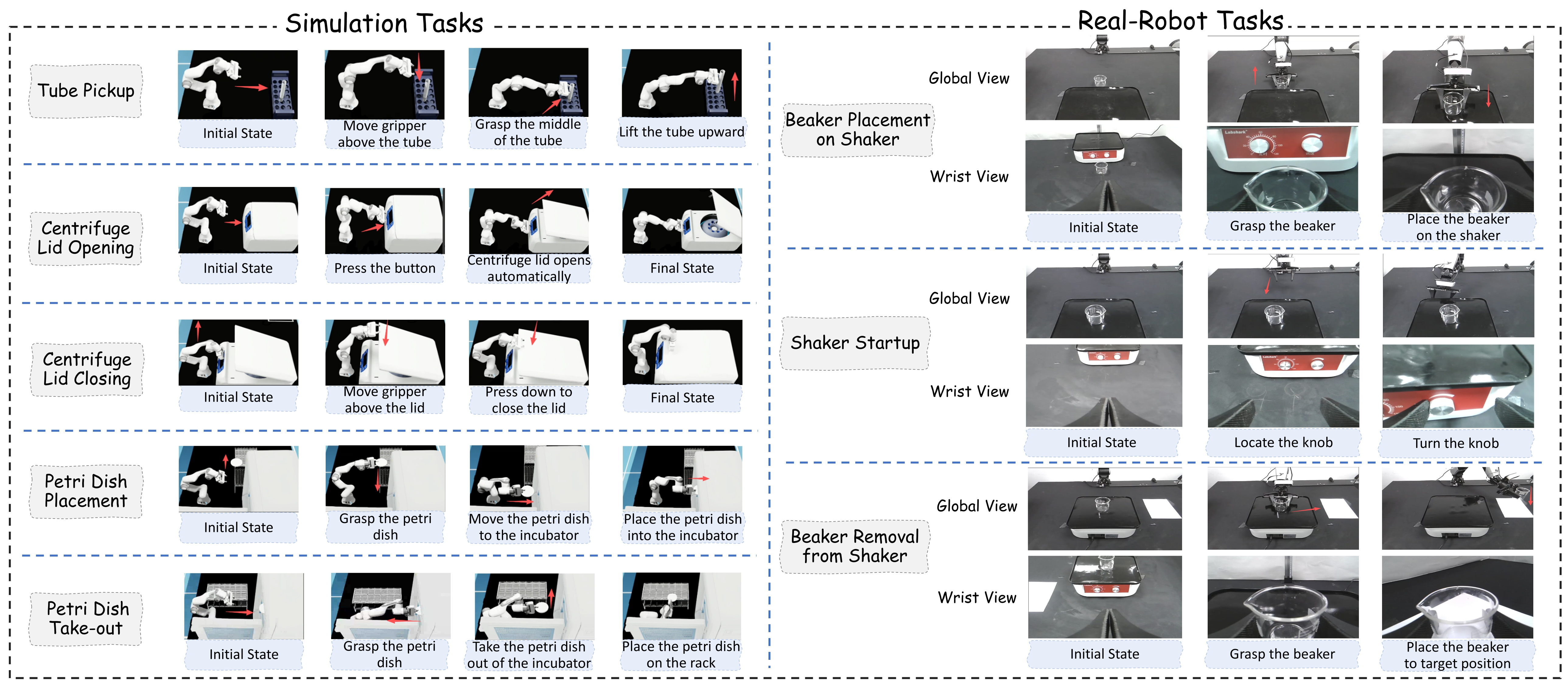}
    \caption{Example embodied execution tasks in simulation and real-robot settings.}
    \label{fig:simulation_tasks}
\end{figure*}

To validate whether the subtasks parsed by ProtoAct can be used for embodied execution, we construct five robotic manipulation tasks in Pipette, an embodied simulation platform for wet-lab robotics~\cite{liu2026pipette}. The left side of Figure~\ref{fig:simulation_tasks} shows their key execution stages.
For each task, 30 trajectories are collected through keyboard teleoperation and augmented by varying scene illumination and execution speed. Additional dataset statistics are listed in the supplementary material.

\subsubsection{Simulation Execution Results}

Using the five simulation task datasets, we train SmolVLA and \(\pi_0\) with the LeRobot framework~\cite{cadene2026lerobot} and evaluate them online in Pipette~\cite{liu2026pipette}. The supplementary material details the evaluation workflow, and the results are reported in Table~\ref{tab:vla_success}.
Overall, SmolVLA achieves higher execution success rates on most tasks, while \(\pi_0\) also completes successful executions. The results show that the subtasks parsed by ProtoAct from biological experimental protocols can guide users without biological expertise in collecting robotic demonstration data and can also serve as language-conditioned inputs for VLA model training and execution, thereby validating their applicability to embodied manipulation tasks.

\subsubsection{Real-Robot Execution Validation}

To validate the subtasks generated by ProtoAct on a physical robot arm, we construct three real-robot tasks. The right side of Figure~\ref{fig:simulation_tasks} shows their key execution stages. Together, they form a continuous workflow. For each task, 100 demonstration trajectories are collected through gamepad teleoperation and used to train SmolVLA and \(\pi_0\) with the LeRobot framework. Both models complete the tasks, with SmolVLA showing more stable execution, consistent with the simulation results. The experiments show that the subtasks generated by ProtoAct can be converted into executable robotic actions in physical environments, validating the practical applicability of the proposed pipeline in connecting protocol understanding with real-world embodied execution.

\section{Conclusion}

This paper introduces ProtoAct, a structured protocol-grounding framework that bridges free-form biological experimental protocols and embodied robotic execution. ProtoAct first produces a human-inspectable representation consisting of monitoring conditions and fine-grained subtasks, and then grounds it into schema-constrained JSON action functions. Through ProtoRAG, RefineChecker, and ActSchema, the framework makes experimental states, action boundaries, parameter details, and procedural dependencies explicit, while supporting both automatic evaluation and downstream robotic execution.
We further construct BioP2E by manually annotating 22 cell-culture protocols, and establish a multi-level benchmark covering action recognition, parameter extraction, and sequence consistency. Experiments across seven large language models demonstrate that ProtoAct can be instantiated with different model backbones. The ablation results confirm the complementary contributions between modules. Simulation and real-robot experiments further show that the generated subtasks can support demonstration collection, VLA model training, and physical robot execution, providing a practical connection between biological protocol understanding and embodied action.

BioP2E is currently limited in scale and biological coverage, RefineChecker depends on experiment-specific information provided by users, and the robotic evaluation focuses on selected manipulation tasks rather than complete closed-loop wet-lab workflows. Future work will expand the protocol domains and action schema, incorporate structured experimental knowledge and autonomous error-recovery mechanisms, and evaluate longer workflows in real laboratory environments.

\bibliography{aaai_references}

\appendix
\section*{Appendix}
\setcounter{subsection}{0}
\renewcommand{\thesubsection}{\Alph{subsection}}
\renewcommand{\thesubsubsection}{\thesubsection.\arabic{subsubsection}}

\begin{figure*}[t]
    \centering
    \includegraphics[width=0.80\textwidth]{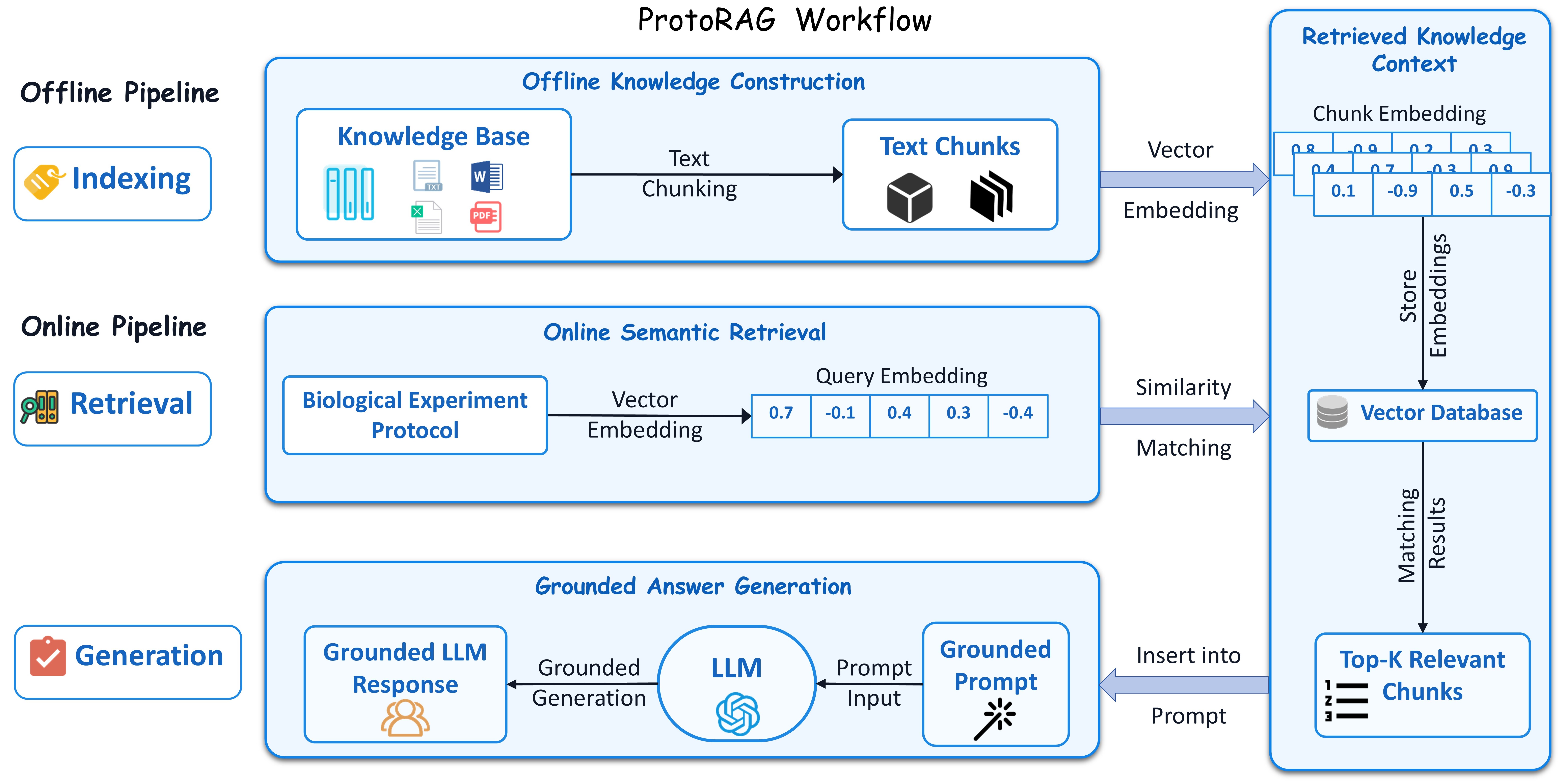}
    \caption{Workflow of the ProtoRAG module.}
    \label{fig:protorag}
\end{figure*}

\begin{figure*}[t]
    \centering
    \includegraphics[width=0.80\textwidth]{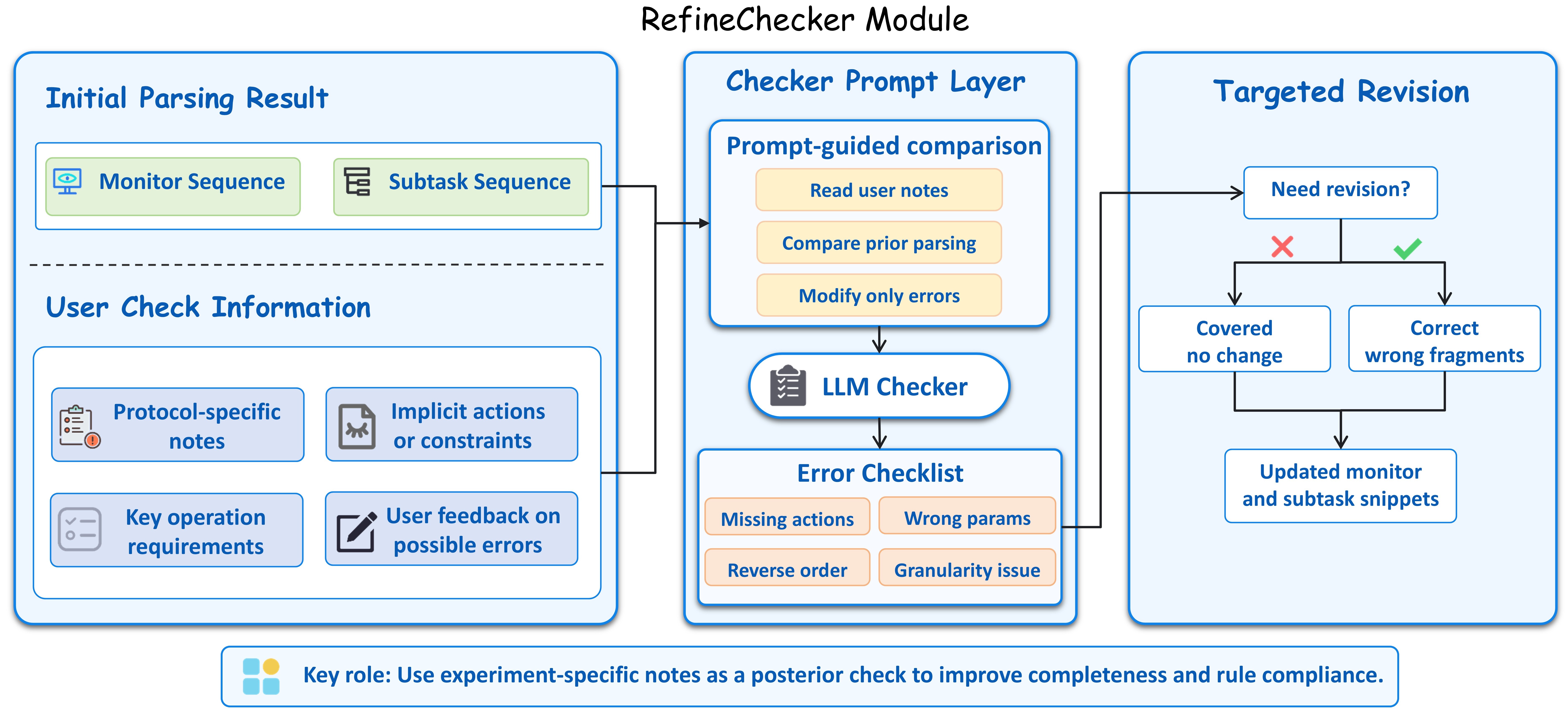}
    \caption{Architecture of the RefineChecker module.}
    \label{fig:refinechecker}
\end{figure*}

\subsection{ProtoRAG Workflow}
\label{app:protorag}

ProtoRAG is used to improve the standardization and completeness of first-stage protocol parsing. Its structure is shown in Figure~\ref{fig:protorag}. Biological experimental protocols are highly domain-dependent: the same type of operation may be expressed differently across texts, and without example-based constraints, large language models may produce overly coarse action decompositions or expressions that do not follow the annotation standard. Therefore, this study builds an external knowledge base from manually annotated experimental protocol fragments. When parsing a new protocol, the system retrieves reference fragments that are semantically similar to the current input and provides them to the large language model together with the parsing rules, helping the model generate monitoring conditions and subtask sequences that better match the task requirements.

ProtoRAG consists of two processes: offline knowledge construction and online retrieval-augmented generation. In the offline stage, manually annotated protocol examples are used as the knowledge base content, where each example contains the original experimental steps and the corresponding standardized parsing result. Long texts are split into chunks, and an embedding model encodes these text fragments into vectors, which are stored in a local Chroma vector database. In the online parsing stage, the raw experimental protocol entered by the user first serves as the query text for the vector retriever. The system returns several document fragments with the highest semantic similarity from the vector database. The retrieved results are then formatted as reference materials and assembled with the role definition, action granularity rules, and other instructions in the system prompt to form the final prompt. When generating the parsing result, the large language model relies not only on its general language understanding ability, but also on the manually annotated results of similar experimental protocols, thereby learning how to organize monitoring conditions, determine subtask granularity, and express different experimental operations in a standardized form.

\subsection{RefineChecker Architecture}
\label{app:refinechecker}

RefineChecker is used to perform posterior checking and targeted revision on the preliminary natural-language parsing result. Its structure is shown in Figure~\ref{fig:refinechecker}. Because biological experimental protocols often contain abbreviations, omissions, and implicit operations, large language models may still misunderstand certain key steps even with the support of reference examples provided by ProtoRAG. RefineChecker is designed to use user-provided experimental notes to check the generated parsing result item by item.

The input of RefineChecker consists of two parts: the monitoring conditions and subtask sequences initially generated by the model, and the checking information supplemented by the user for the current experiment. This information includes key operation flows in the protocol, experimental constraints, and implicit actions. Based on the user-provided checking information, the model compares it with the historical parsing result. If the original parsing result already covers the relevant notes, no modification is needed. If omissions or errors are confirmed, the erroneous fragment is reanalyzed, and the revised monitoring conditions and subtask fragments are output.

\subsection{BioP2E Annotation and Dataset Statistics}
\label{app:annotation}

\subsubsection{Annotation Procedure}

\begin{figure*}[t]
    \centering
    \includegraphics[width=0.88\textwidth]{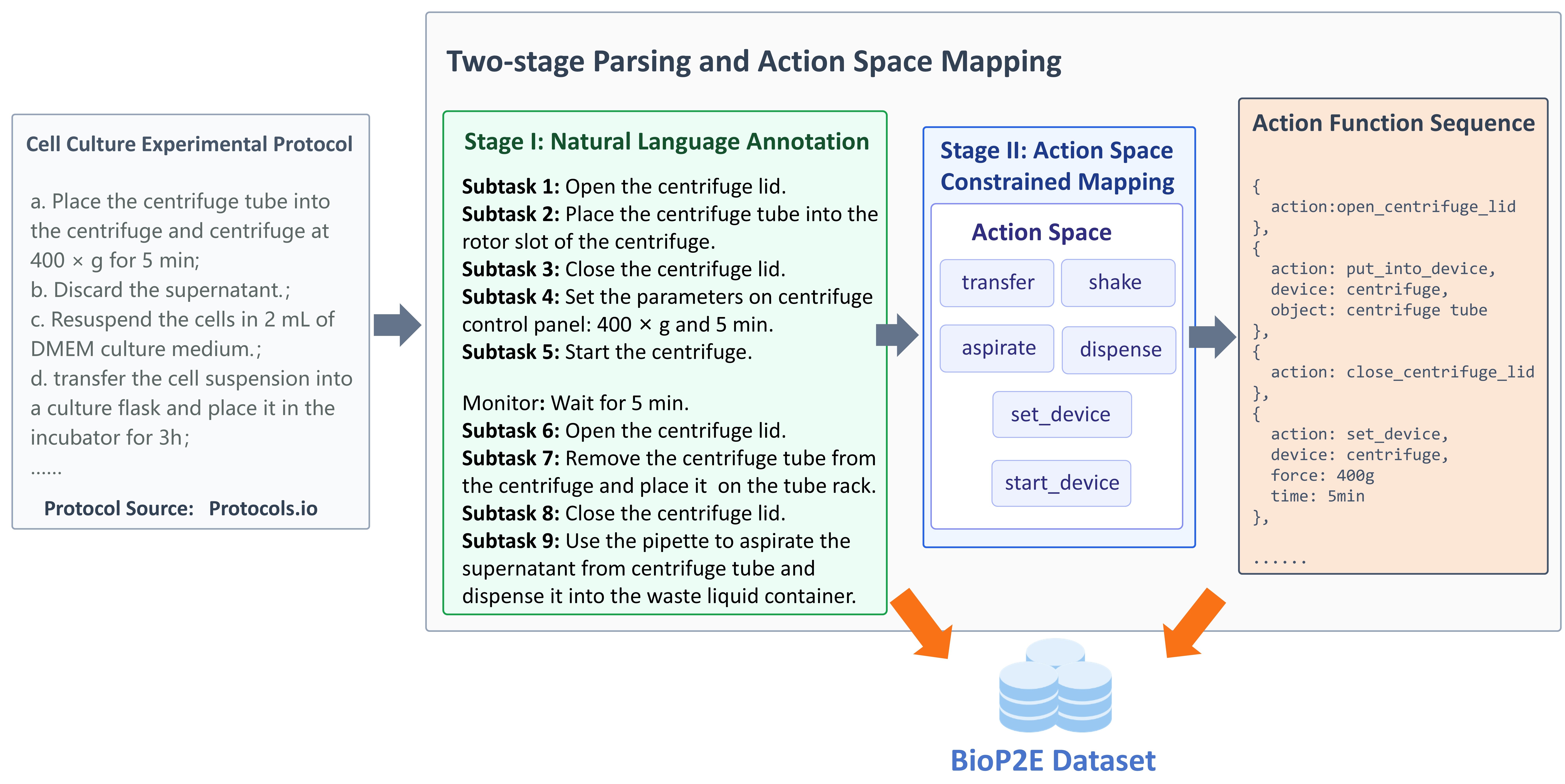}
    \caption{Annotation procedure of the BioP2E dataset.}
    \label{fig:annotation_procedure}
\end{figure*}

The annotation process of BioP2E consists of two stages, as shown in Figure~\ref{fig:annotation_procedure}. In the first stage, the raw protocol is parsed into a natural-language monitoring conditions and subtask sequences. The monitoring sequence describes experimental states or conditions that trigger subsequent operations, such as the completion of device operation, waiting for a specified duration, observing a particular cell state, or confirming that experimental resources are ready. The subtask sequence describes individual actions executed by the robotic arm, such as aspirating liquid, placing a container, putting an object into a centrifuge, setting device parameters, or starting a device. The main goal of this stage is to decompose complex experimental procedures into operation units with clear execution boundaries and experimental semantics, thereby providing a natural-language task representation for downstream robotic understanding and execution.

In the second stage, the monitoring conditions and subtask sequences are further mapped into JSON action function sequences. This stage mainly supports automated evaluation. By converting both manually annotated results and large-language-model-generated parsing results into structured action function sequences, the framework can compute metrics such as action matching, parameter extraction, parameter-value similarity, and action-order consistency. To avoid incomparable outputs caused by freely generated action names, this paper predefines an action space for biological experimental operations. As summarized in Table~\ref{tab:action_space}, the action space contains four action types: general manipulation actions, liquid-handling actions, device-operation actions, and coarse-grained biological actions. Each action function has a fixed action name and parameter fields, ensuring consistency in both semantic expression and evaluation computation.

\begin{table*}[!t]
\centering
\small
\setlength{\tabcolsep}{2.5pt}
\begin{tabular}{@{}L{0.17\textwidth}L{0.22\textwidth}L{0.38\textwidth}C{0.065\textwidth}C{0.065\textwidth}@{}}
\toprule
Category & Example Action & Required Params and Meanings & \#Action & \#Param. \\
\midrule
General manipulation actions & pick\_and\_place & object: item to move, e.g., culture dish; place: target location, e.g., workbench & 4 & 8 \\
& shake & object: item to shake, e.g., centrifuge tube & & \\
\addlinespace[3pt]
Liquid-handling actions & aspirate & object: liquid to aspirate, e.g., DMEM culture medium; amount: volume, e.g., 6 mL & 5 & 8 \\
& dispense & place: target container or location, e.g., culture dish; amount: volume, e.g., 6 mL & & \\
\addlinespace[3pt]
Device-operation actions & put\_into\_centrifuge & object: item to load, e.g., centrifuge tube & 10 & 15 \\
& set\_device & device: target device, e.g., centrifuge; speed: device setting, e.g., 800 r/min; time: duration, e.g., 10 min & & \\
& take\_out\_centrifuge & object: item to unload, e.g., centrifuge tube & & \\
\addlinespace[3pt]
Coarse-grained biological actions & resuspension & object: sample or container to resuspend, e.g., centrifuge tube & 11 & 25 \\
& cell\_seed & source: source container, e.g., centrifuge tube; place: destination container, e.g., culture flask; volume: operation volume, e.g., 2 mL & & \\
\bottomrule
\end{tabular}
\caption{Action-space statistics and representative action schemas.}
\label{tab:action_space}
\end{table*}

\subsubsection{Dataset Scale and Distribution}
\label{app:biop2e_statistics}

\begin{figure*}[!t]
    \centering
    \includegraphics[width=0.80\textwidth]{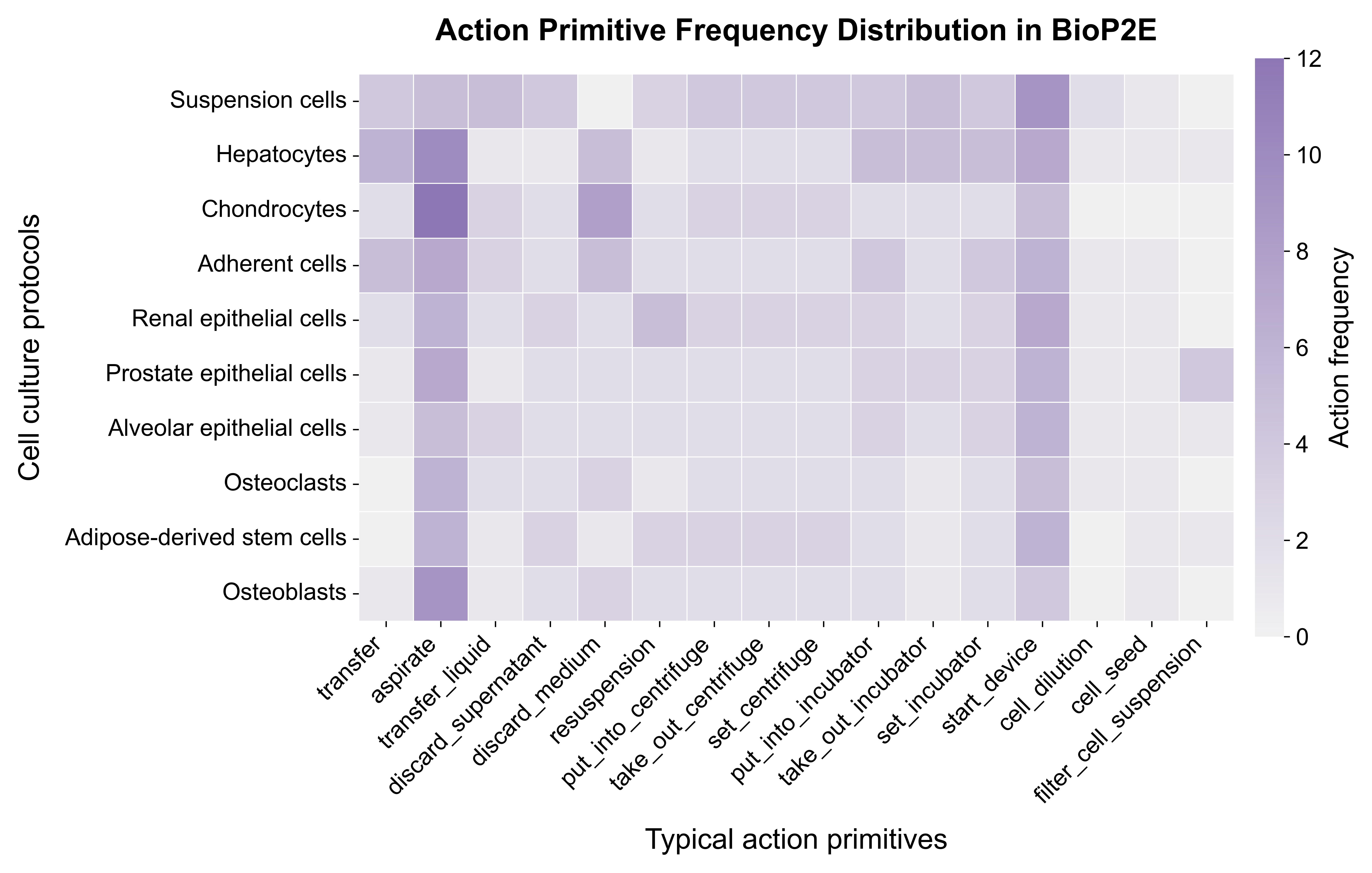}
    \caption{Action primitive frequency heatmap for selected BioP2E cell culture protocols.}
    \label{fig:biop2e_action_frequency}
    \vspace{0.2em}
    \includegraphics[width=0.80\textwidth]{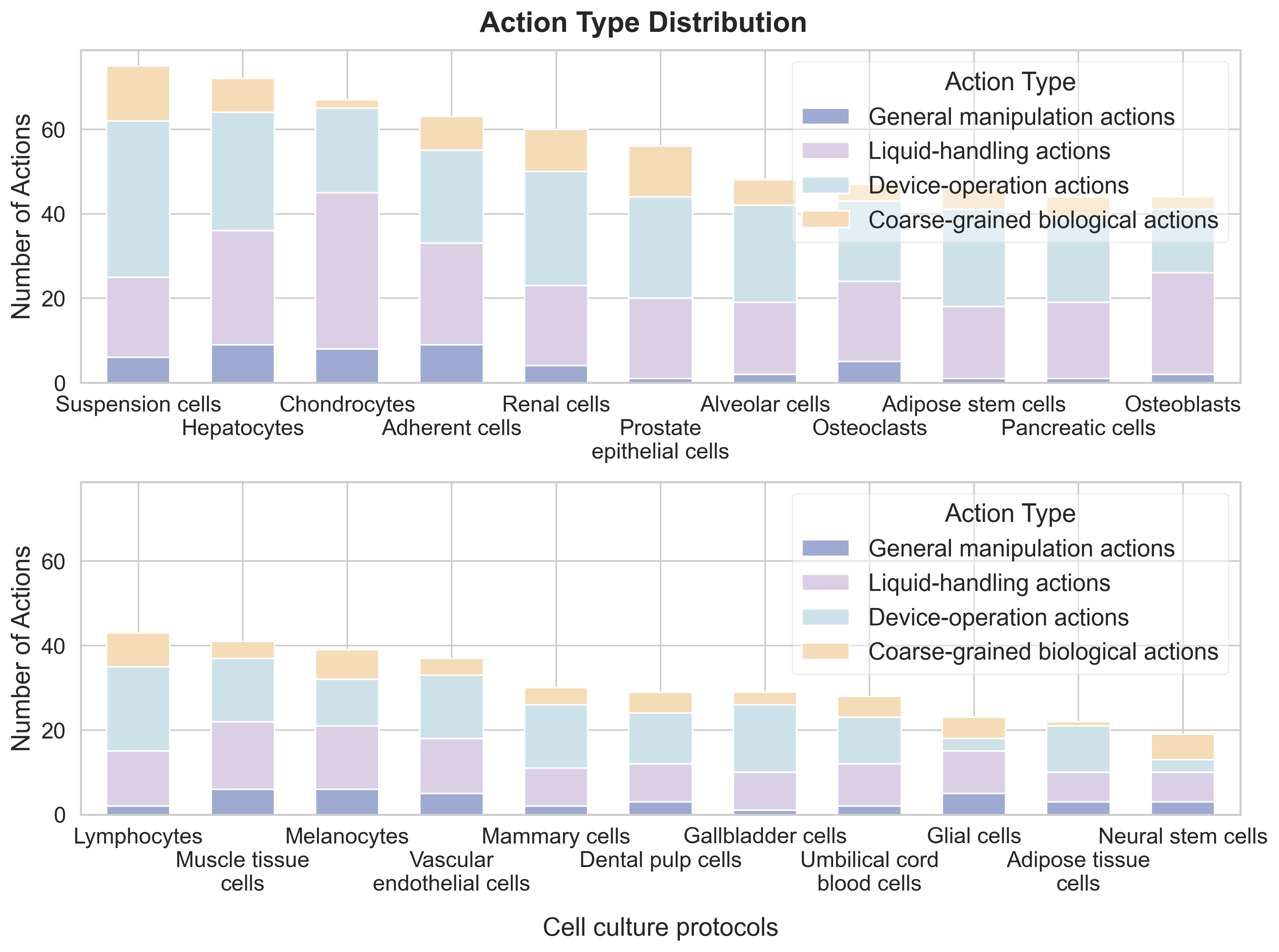}
    \caption{Action type distribution across BioP2E cell culture protocols.}
    \label{fig:biop2e_action_type_distribution}
\end{figure*}

To present the scale and distributional characteristics of BioP2E, we conduct statistical analyses of the dataset from multiple perspectives. Figure~\ref{fig:biop2e_action_frequency} shows the distribution of different action primitives used in selected BioP2E protocols. Actions such as \actionname{aspirate}, \actionname{transfer}, \actionname{discard_medium}, \actionname{start_device}, and \actionname{put_into_incubator} appear frequently in most protocols, indicating that liquid handling, transfer operations, and device operations are typical actions in cell culture experiments. Overall, this heatmap shows that BioP2E contains diverse action primitives and clear inter-protocol variation, providing a fine-grained data basis for evaluating the action-mapping ability of large language models and constructing downstream embodied robotic tasks.

Figure~\ref{fig:biop2e_action_type_distribution} shows the distribution of action types across BioP2E protocols. The figure indicates that liquid-handling and device-operation actions account for relatively large proportions in most protocols, while some protocols also contain a certain proportion of coarse-grained biological actions. This demonstrates that the manual annotations in BioP2E cover not only basic experimental operations but also key procedural stages with biological experimental semantics.

\begin{figure}[t]
    \centering
    \includegraphics[width=\columnwidth]{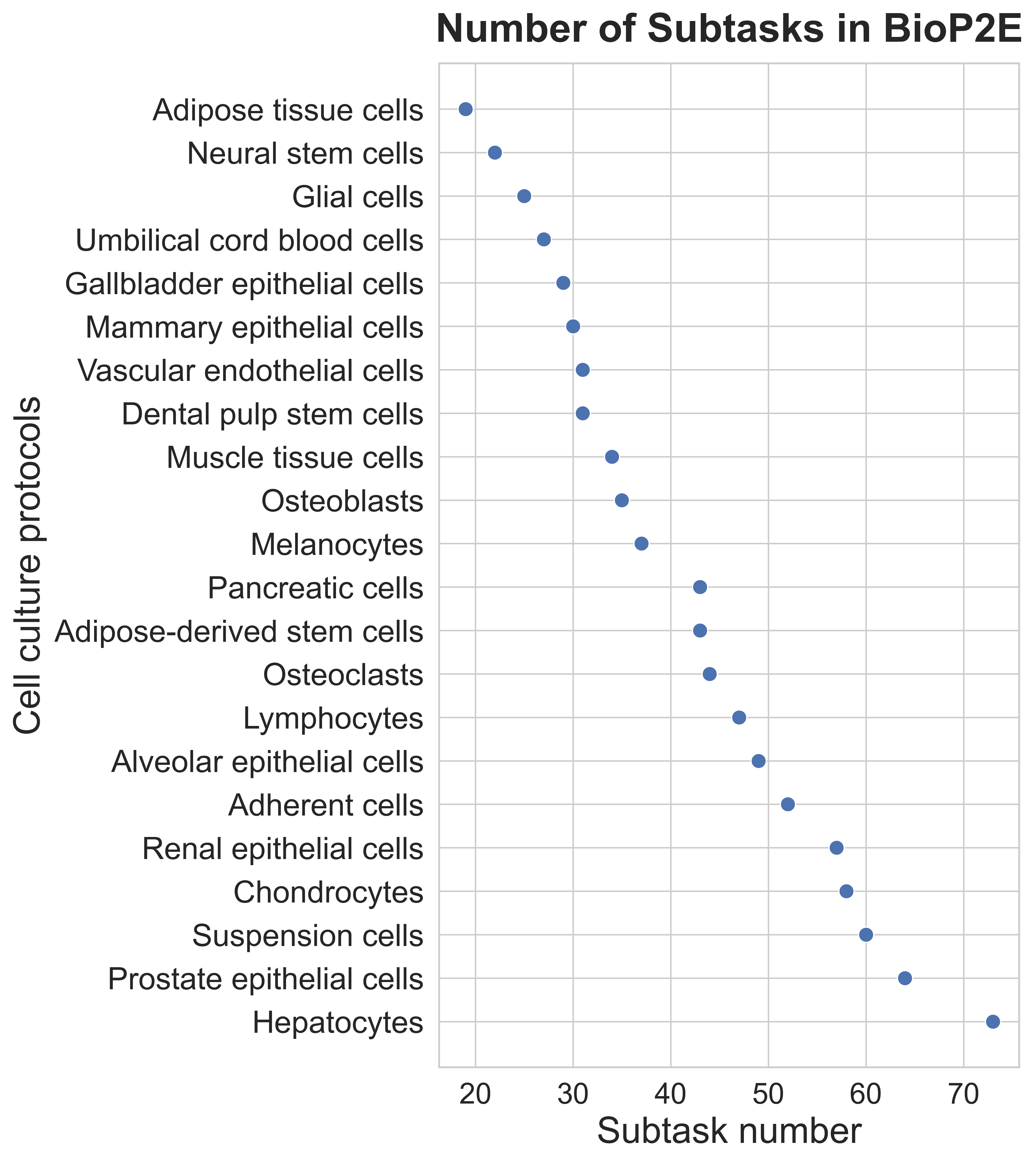}
    \caption{Distribution of manually annotated subtask counts across BioP2E cell culture protocols. Each point denotes one protocol; the x-axis indicates the number of natural-language subtasks, and the y-axis sorts protocols in ascending order by subtask count.}
    \label{fig:biop2e_subtask_count}
\end{figure}

Figure~\ref{fig:biop2e_subtask_count} shows the distribution of subtask counts across BioP2E protocols. Simpler protocols are parsed into around 20 subtasks, whereas more complex protocols can be decomposed into more than 70 subtasks. This indicates that BioP2E contains many fine-grained and executable experimental operation units.

These statistics show that although BioP2E contains a limited number of raw protocols, each protocol is decomposed into a large number of executable subtasks, resulting in densely structured annotations. Therefore, the core value of BioP2E lies in expanding a small set of professional experimental protocols into multi-level annotated data containing monitoring conditions, subtask sequences, and structured action functions, thereby providing a unified data basis for large-language-model-based protocol parsing, automated evaluation, and embodied robotic task validation.

\begin{figure}[t]
    \centering
    \includegraphics[width=\columnwidth]{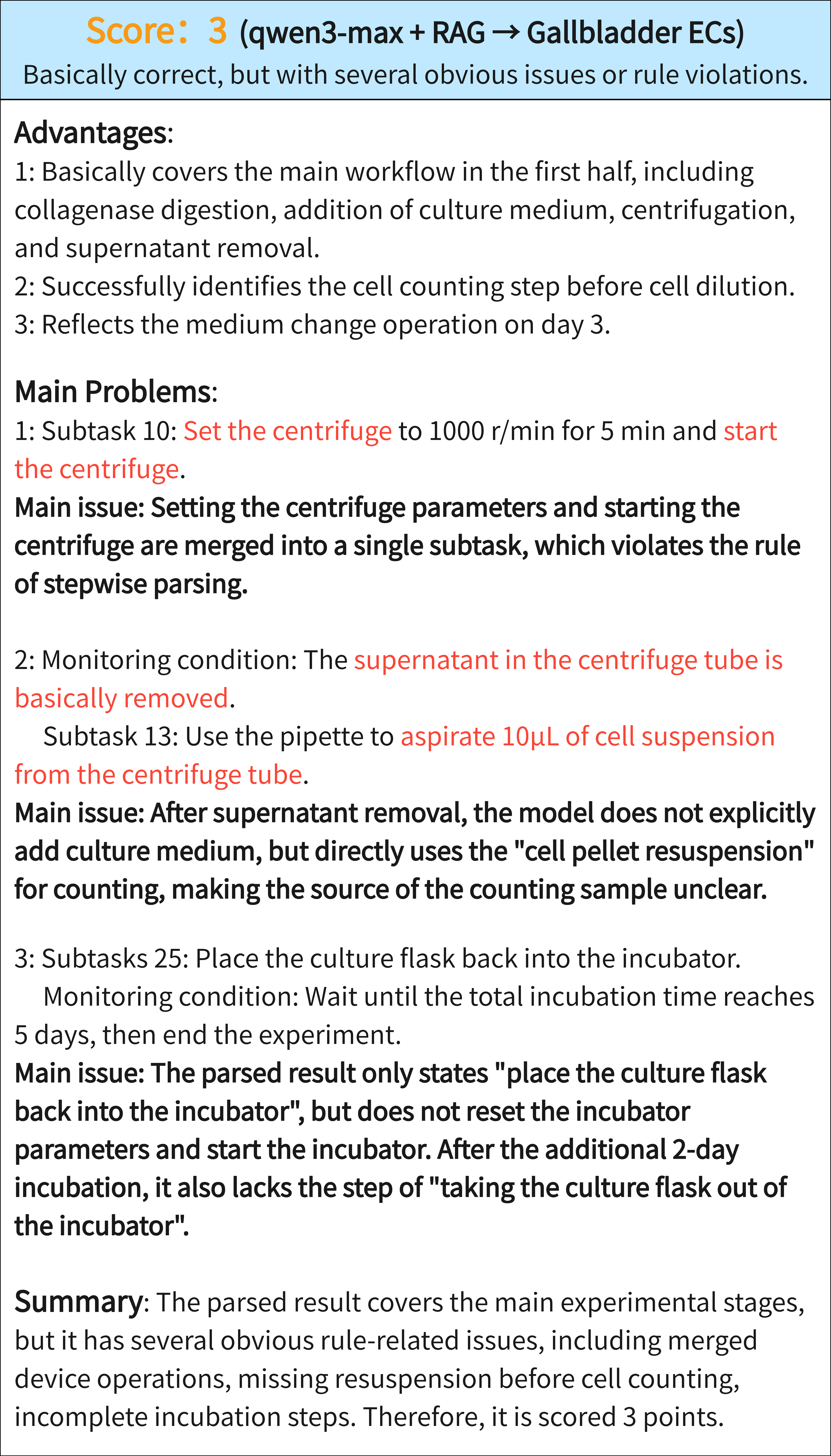}
    \caption{Output example of GPT-based parsing quality evaluation.}
    \label{fig:gpt_eval_example}
\end{figure}

\subsection{GPT Evaluation Case}
\label{app:gpt_eval_example}

Figure~\ref{fig:gpt_eval_example} shows an example output produced by the GPT evaluator. The output consists of four parts. Score denotes the final rating, ranging from 1 to 5, with increments of 0.5 allowed for intermediate levels.
Advantages summarizes the experimental stages and key operations correctly covered by the parsing result. Main Problems identifies issues such as rule violations, missing actions and parameters, incorrect procedural order, or inappropriate operation granularity. Summary provides an overall assessment of the parsing quality and the rationale behind the assigned score. The meanings of different scores are as follows: A score of 5 indicates that the parsed result is complete, accurate, and strictly follows the parsing rules. Scores of 4 or 4.5 indicate that the overall procedure is correct, with only minor omissions or limited detail level issues. Scores of 3 or 3.5 indicate that the procedure is generally correct but contains several noticeable errors or some rule violations. Scores of 2 or 2.5 indicate that the procedure has substantial structural or logical problems, while still retaining partially usable content. Scores of 1 or 1.5 indicate that the parsed result contains severe overall errors, disordered structure, clear rule violations, or omissions of key steps.

\subsection{Ablation Results}
\label{app:ablation}

Tables~\ref{tab:abl_monitor_type} and~\ref{tab:abl_action_type} provide the complete ablation results for ProtoRAG, ActSchema, and RefineChecker. Since the main text already discusses the key trends, this section only reports the complete tables and briefly summarizes the GPT-based ablation results.

\begin{figure*}[t]
    \centering
    \includegraphics[width=0.88\textwidth]{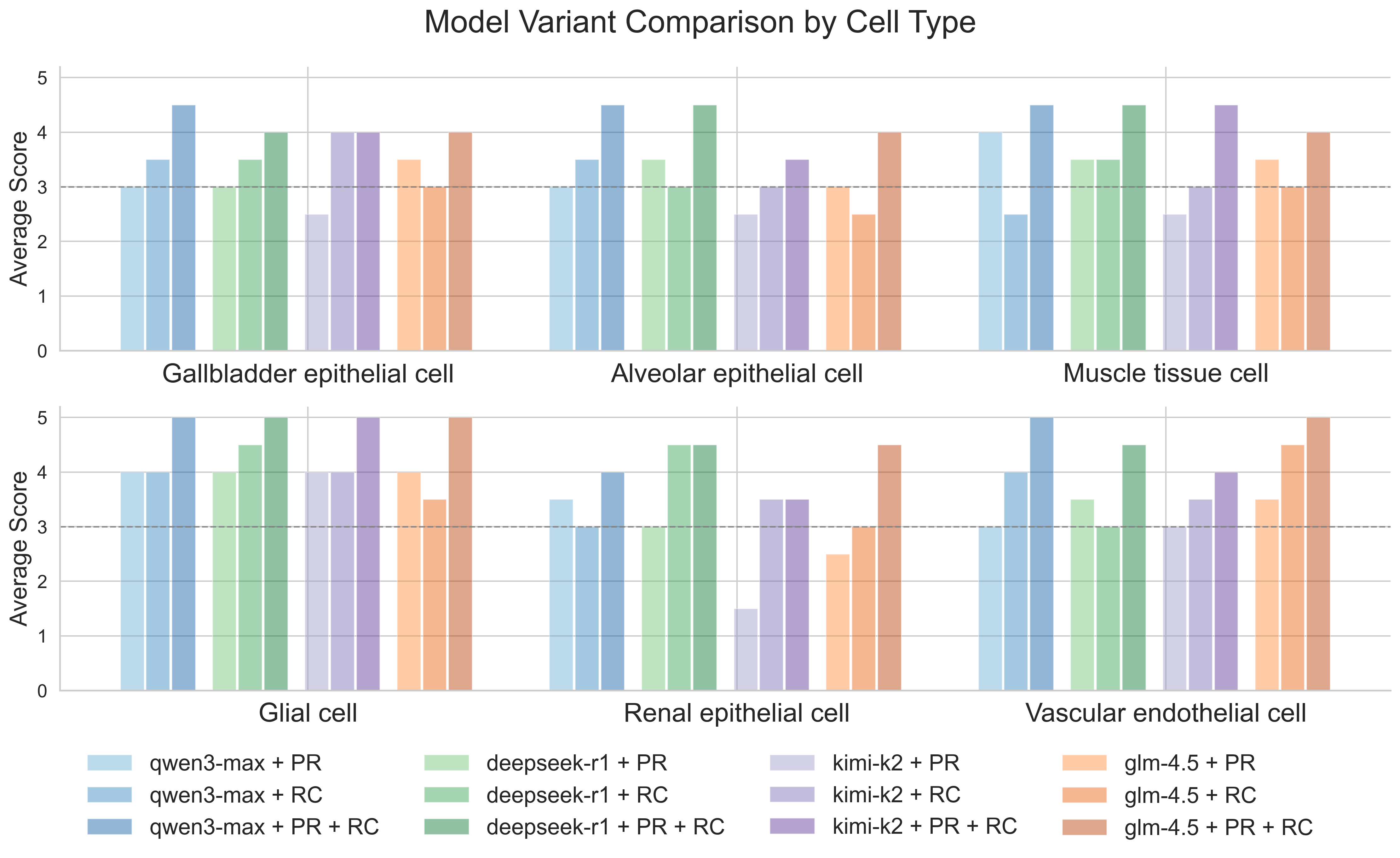}
    \caption{GPT score comparison for model-module combinations across six cell culture protocol parsing tasks. PR and RC denote ProtoRAG and RefineChecker, respectively.}
    \label{fig:gpt_score_comparison}
\end{figure*}

Figure~\ref{fig:gpt_score_comparison} presents the overall scores assigned by the GPT evaluator to the natural-language parsing results generated by different models and module combinations. The evaluated outputs are the monitoring conditions and subtask sequences produced in the first parsing stage. Therefore, this section mainly compares the effects of ProtoRAG and RefineChecker on parsing quality. As shown in the figure, models using only ProtoRAG obtain relatively lower scores. 
The reason is that the retrieved reference examples cannot fully prevent missing actions, incomplete procedures, or rule violations. After introducing RefineChecker, most models achieve higher scores across different cell protocol parsing tasks, indicating that the posterior checking mechanism helps models identify and revise key omissions and procedural-order errors.

Overall, the GPT-evaluator results verify the complementary roles of ProtoRAG and RefineChecker in improving protocol parsing quality. ProtoRAG provides manually annotated examples and relevant contextual information, making it easier for models to learn the standard parsing format, action decomposition granularity, and expression style of experimental procedures. RefineChecker further reviews the preliminary parsing results according to predefined rules and experiment-specific notes, thereby reducing missing actions and procedural misalignment. 
When the two modules are combined, the resulting parses are of higher quality, which is consistent with the conclusions drawn from the ablation experiments.

\subsection{Simulation Dataset Details}
\label{app:simulation_dataset}

This study constructs five robotic-arm manipulation tasks in the Isaac Sim simulation platform, including picking up the tube from the tube rack, opening the centrifuge lid, closing the centrifuge lid, taking a culture dish out of the incubator, and placing a culture dish into the incubator. These tasks are all derived from typical subtasks in the ProtoAct parsing results.

During data collection, each task is demonstrated by controlling the robotic arm through keyboard teleoperation. The robot joint states, visual observations, and language instructions are saved as HDF5 data. Table~\ref{tab:simulation_dataset} shows the dataset statistics for the five simulation tasks. For each task, 30 trajectories are collected through keyboard teleoperation, and the collected data are then augmented offline to improve training-data diversity. The augmentation process mainly expands the data from visual aspects, including adjusting scene lighting intensity and changing trajectory execution speed, thereby improving the adaptability of VLA models to different observation conditions and action rhythms.

It should be noted that the numbers of augmented samples are not exactly the same across tasks because each augmented trajectory is automatically checked according to the success criterion of the corresponding task. If an augmented trajectory fails to satisfy the task-success metric, the augmented sample is automatically discarded. This filtering mechanism ensures that the augmented data maintain high task validity, providing a reliable data basis for subsequent VLA model training and simulation evaluation.

\begin{table}[t]
\centering
\small
\setlength{\tabcolsep}{2pt}
\begin{tabular}{@{}L{0.30\columnwidth}L{0.43\columnwidth}cc@{}}
\toprule
Task & Instruction & Orig. & Aug. \\
\midrule
Tube pickup & pick\_up\_the\_tube & 30 & 118 \\
Lid opening & open\_centrifuge\_lid & 30 & 119 \\
Lid closing & close\_centrifuge\_lid & 30 & 120 \\
Dish take-out & take\_out\_dish & 30 & 111 \\
Dish placement & place\_dish & 30 & 118 \\
\bottomrule
\end{tabular}
\caption{Dataset statistics of five robotic manipulation tasks in the simulation platform.}
\label{tab:simulation_dataset}
\end{table}

\subsection{VLA Evaluation Workflow}
\label{app:vla_eval_workflow}

\begin{figure*}[t]
    \centering
    \includegraphics[width=0.98\textwidth]{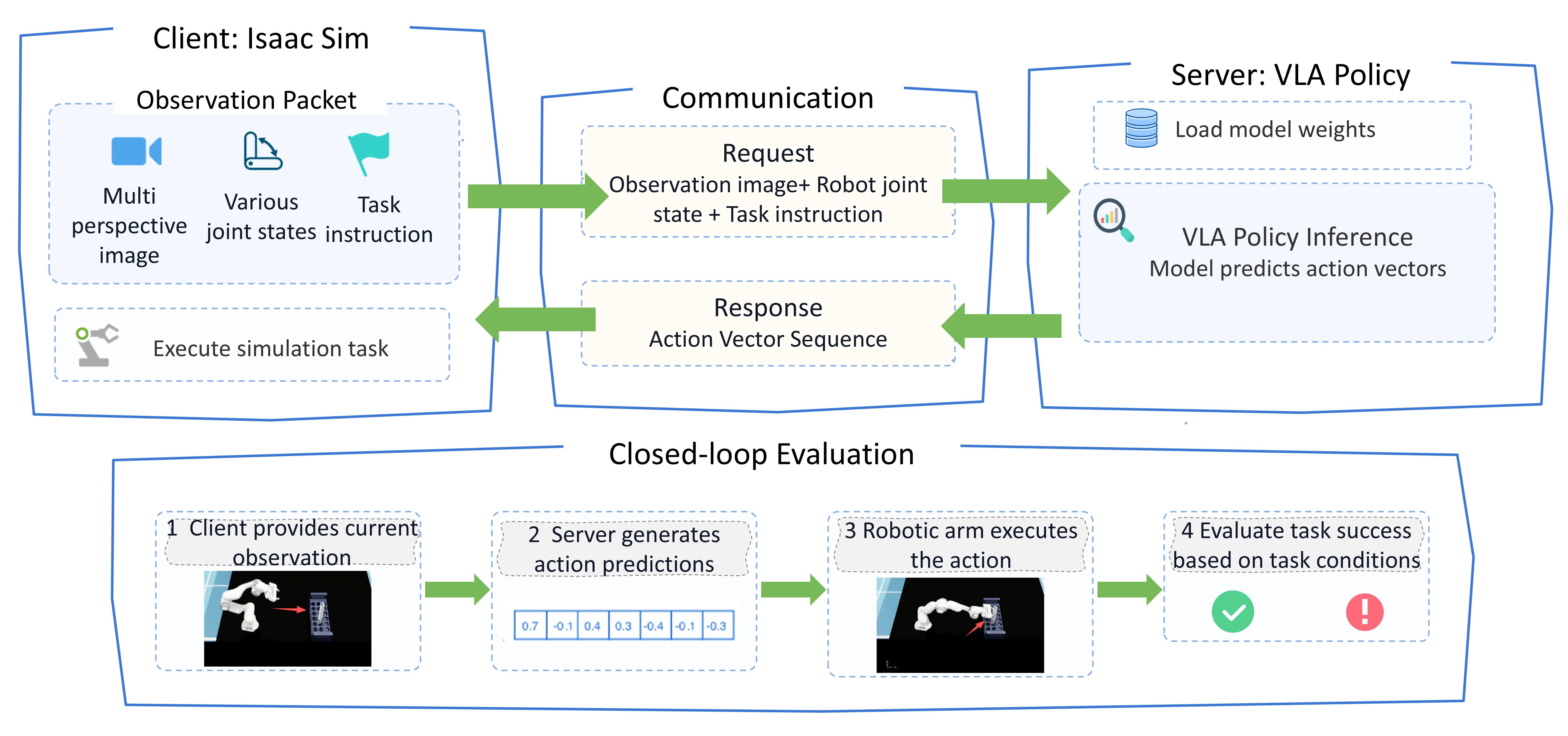}
    \caption{Online evaluation workflow of VLA models in the Isaac Sim simulation platform.}
    \label{fig:vla_eval_workflow}
\end{figure*}

After training two VLA models on the collected data, we evaluate their performance in the Isaac Sim simulation platform. As shown in Figure~\ref{fig:vla_eval_workflow}, the evaluation process adopts a separated server-client structure. The server loads the trained VLA policy model and receives observation inputs from the simulation environment. The client executes tasks in Isaac Sim, collects multi-view visual observations and robot states, and sends them to the server for policy inference. After receiving the predicted action from the server, the client applies the action to the robotic arm and records the execution result according to the task-specific success condition. This workflow simulates the closed-loop input-output process of VLA model execution.

\begin{table*}[t]
\centering
\small
\setlength{\tabcolsep}{0.7mm}
\begin{tabular}{lccccccccccc}
\toprule
Model & PR & AS & RC & \multicolumn{3}{c}{Monitor Type} & \multicolumn{5}{c}{Monitor Param} \\
\cmidrule(lr){5-7}\cmidrule(lr){8-12}
 &  &  &  & Precision & Recall & F1 & Precision & Recall & F1 & BLEU & SciBERT \\
\midrule
qwen3-max & \(\surd\) & \(\surd\) & \(\surd\) & \textbf{0.98{\(\pm\)}0.05} & \textbf{0.88{\(\pm\)}0.07} & \textbf{0.93{\(\pm\)}0.04} & \textbf{0.98{\(\pm\)}0.05} & \textbf{0.89{\(\pm\)}0.07} & \textbf{0.93{\(\pm\)}0.04} & \textbf{0.77{\(\pm\)}0.11} & \textbf{0.96{\(\pm\)}0.02} \\
qwen3-max & \(\times\) & \(\surd\) & \(\surd\) & \underline{0.97{\(\pm\)}0.08} & 0.79{\(\pm\)}0.10 & 0.87{\(\pm\)}0.10 & \underline{0.97{\(\pm\)}0.07} & 0.81{\(\pm\)}0.12 & 0.89{\(\pm\)}0.10 & \underline{0.76{\(\pm\)}0.11} & \underline{0.96{\(\pm\)}0.03} \\
qwen3-max & \(\surd\) & \(\times\) & \(\surd\) & 0.96{\(\pm\)}0.07 & \underline{0.85{\(\pm\)}0.09} & \underline{0.90{\(\pm\)}0.06} & 0.96{\(\pm\)}0.07 & \underline{0.87{\(\pm\)}0.08} & \underline{0.91{\(\pm\)}0.06} & 0.75{\(\pm\)}0.09 & 0.96{\(\pm\)}0.02 \\
qwen3-max & \(\surd\) & \(\surd\) & \(\times\) & 0.96{\(\pm\)}0.07 & 0.82{\(\pm\)}0.10 & 0.89{\(\pm\)}0.06 & 0.96{\(\pm\)}0.06 & 0.83{\(\pm\)}0.09 & 0.89{\(\pm\)}0.05 & 0.73{\(\pm\)}0.09 & 0.95{\(\pm\)}0.02 \\
\bottomrule
\end{tabular}
\caption{Monitor condition parsing results in the ablation experiment. PR, AS, and RC denote ProtoRAG, ActSchema, and RefineChecker, respectively.}
\label{tab:abl_monitor_type}
\label{tab:abl_monitor_param}
\end{table*}

\begin{table*}[t]
\centering
\small
\setlength{\tabcolsep}{0.7mm}
\begin{tabular}{lcccccccccccc}
\toprule
Model & PR & AS & RC & \multicolumn{3}{c}{Action Primitive} & \multicolumn{5}{c}{Action Param} & Action Order \\
\cmidrule(lr){5-7}\cmidrule(lr){8-12}\cmidrule(lr){13-13}
 &  &  &  & Precision & Recall & F1 & Precision & Recall & F1 & BLEU & SciBERT & \(L_{\mathrm{dn}}\) \\
\midrule
qwen3-max & \(\surd\) & \(\surd\) & \(\surd\) & \textbf{0.97{\(\pm\)}0.03} & \textbf{0.97{\(\pm\)}0.03} & \textbf{0.97{\(\pm\)}0.03} & \textbf{0.98{\(\pm\)}0.03} & \textbf{0.96{\(\pm\)}0.04} & \textbf{0.97{\(\pm\)}0.03} & \textbf{0.88{\(\pm\)}0.07} & \textbf{0.98{\(\pm\)}0.01} & \textbf{0.05{\(\pm\)}0.04} \\
qwen3-max & \(\times\) & \(\surd\) & \(\surd\) & \underline{0.96{\(\pm\)}0.04} & \underline{0.96{\(\pm\)}0.03} & \underline{0.96{\(\pm\)}0.03} & \underline{0.97{\(\pm\)}0.04} & \underline{0.96{\(\pm\)}0.05} & \underline{0.97{\(\pm\)}0.04} & \underline{0.85{\(\pm\)}0.09} & 0.98{\(\pm\)}0.02 & \underline{0.07{\(\pm\)}0.05} \\
qwen3-max & \(\surd\) & \(\times\) & \(\surd\) & 0.91{\(\pm\)}0.04 & 0.90{\(\pm\)}0.05 & 0.91{\(\pm\)}0.04 & 0.91{\(\pm\)}0.06 & 0.90{\(\pm\)}0.07 & 0.90{\(\pm\)}0.06 & 0.84{\(\pm\)}0.07 & \underline{0.98{\(\pm\)}0.01} & 0.14{\(\pm\)}0.06 \\
qwen3-max & \(\surd\) & \(\surd\) & \(\times\) & 0.95{\(\pm\)}0.04 & 0.90{\(\pm\)}0.08 & 0.92{\(\pm\)}0.05 & 0.97{\(\pm\)}0.03 & 0.91{\(\pm\)}0.07 & 0.94{\(\pm\)}0.04 & 0.84{\(\pm\)}0.10 & 0.98{\(\pm\)}0.02 & 0.15{\(\pm\)}0.07 \\
\bottomrule
\end{tabular}
\caption{Action sequence parsing results in the ablation experiment. PR, AS, and RC denote ProtoRAG, ActSchema, and RefineChecker, respectively.}
\label{tab:abl_action_type}
\label{tab:abl_action_param}
\end{table*}

\end{document}